%% file: iclr2027_conference.tex
\documentclass{article} % For LaTeX2e
\usepackage{iclr2027_conference,times}

\input{math_commands.tex}

\usepackage{hyperref}
\usepackage{url}
\usepackage{booktabs}
\usepackage{graphicx}
\usepackage{amssymb}
\usepackage{multirow}
\usepackage{xcolor}
\usepackage{colortbl}
\usepackage{tabularx}
\usepackage{array}
\usepackage{pifont}
\usepackage{caption}
\usepackage{mdframed}
\usepackage{xcolor}
\usepackage[table]{xcolor}
\usepackage{booktabs}
\usepackage{makecell}
\usepackage{multirow}
\usepackage{pifont}
\usepackage{caption}

\definecolor{headergray}{RGB}{247,247,247}
\definecolor{baselinegray}{RGB}{249,249,249}
\definecolor{backbonegray}{RGB}{244,246,248}

\definecolor{groupAHead}{RGB}{236,243,249}
\definecolor{groupABody}{RGB}{247,250,253}

\definecolor{groupBHead}{RGB}{247,241,229}
\definecolor{groupBBody}{RGB}{252,249,242}

\definecolor{groupCHead}{RGB}{237,246,236}
\definecolor{groupCBody}{RGB}{248,252,247}

\definecolor{groupDHead}{RGB}{243,237,248}
\definecolor{groupDBody}{RGB}{250,247,252}

\definecolor{groupEHead}{RGB}{250,239,228}
\definecolor{groupEBody}{RGB}{253,248,241}

\definecolor{groupFHead}{RGB}{237,244,246}
\definecolor{groupFBody}{RGB}{248,251,252}

\definecolor{judgeA}{RGB}{247,250,253} % light blue
\definecolor{judgeB}{RGB}{250,250,250} % light gray
\definecolor{judgeC}{RGB}{250,247,252} % light purple

\definecolor{ablationA}{RGB}{247,250,253} % light blue
\definecolor{ablationB}{RGB}{252,249,242} % light beige
\definecolor{ablationC}{RGB}{248,252,247} % light green

\definecolor{checkgreen}{RGB}{20,145,70}
\definecolor{crossred}{RGB}{200,65,65}
\definecolor{circleorange}{RGB}{205,145,45}

\newcommand{\cmark}{\textcolor{checkgreen}{\ding{51}}}

\newcolumntype{Y}{>{\centering\arraybackslash}X}
\makeatletter
\providecommand{\makecell}{\@ifnextchar[{\makecell@opt}{\makecell@plain}}
\newcommand{\makecell@opt}[2][]{\shortstack[#1]{#2}}
\newcommand{\makecell@plain}[1]{\shortstack{#1}}
\makeatother
\providecommand{\Cref}[1]{\autoref{#1}}

\definecolor{paperrow}{RGB}{242,242,242}
\definecolor{paperhighlight}{RGB}{231,240,247}
\newcommand{\itbf}[1]{\textit{\textbf{#1}}}

\newcommand{\takeaway}[1]{%
  \par\smallskip
  \begingroup\setlength{\fboxsep}{5pt}%
  \noindent\colorbox{paperhighlight}{%
    \parbox{\dimexpr\columnwidth-2\fboxsep\relax}{%
      \textbf{\textit{Takeaway:}} \textit{#1}}}%
  \endgroup
  \smallskip\par}

\title{What Really Works for Universal Multimodal Embeddings? A Controlled Study of Contrastive Training Strategies}

\author{
\textbf{Duo Li\textsuperscript{1},
Haojun Pan\textsuperscript{2},
Zuhao Yang\textsuperscript{1},
Xintong Wang\textsuperscript{2,3,4},}\\
\textbf{Chengfu Huo\textsuperscript{2},
Shijian Lu\textsuperscript{1,\dag},
Liang Ding\textsuperscript{2,\dag}}\\[1ex]
\textsuperscript{1}CCDS, Nanyang Technological University
\qquad
\textsuperscript{2}Alibaba Group\\
\textsuperscript{3}Universität Hamburg
\qquad
\textsuperscript{4}Sichuan University\\[1ex]
\texttt{\{duo001, Shijian.Lu\}@ntu.edu.sg}\\
\texttt{\{haojun.phj, zuorui.dl\}@alibaba-inc.com}
}

\iclrfinalcopy % Uncomment for camera-ready version, but NOT for submission.
\begin{document}

\raggedbottom

\maketitle

\begingroup
\renewcommand{\thefootnote}{\fnsymbol{footnote}}
\footnotetext[2]{Corresponding authors.}
\endgroup

\input{sec/0_abstract}

\input{sec/1_intro}
\input{sec/2_related_work}

\input{sec/3_exp}
\input{sec/4_analysis}
\input{sec/5_conclusion}

\subsection*{AI use statement}

In this work, we used generative AI tools to assist with polishing the manuscript for clarity and readability. Additionally, we used generative AI tools for editing a research paper to improve readability. All AI-assisted content was critically reviewed by the authors. We take responsibility for the final content of this work, including text, claims or artifacts produced with the aid of generative AI.

\bibliography{iclr2027_conference}
\bibliographystyle{iclr2027_conference}

\clearpage
\appendix

\counterwithin{table}{section}
\counterwithin{figure}{section}
\counterwithin{equation}{section}

\input{sec/6_appendix}

\end{document}

%% file: math_commands.tex
\usepackage{amsmath,amsfonts,bm}

\def\eqref#1{equation~\ref{#1}}
\def\1{\bm{1}}

\DeclareMathAlphabet{\mathsfit}{\encodingdefault}{\sfdefault}{m}{sl}
\SetMathAlphabet{\mathsfit}{bold}{\encodingdefault}{\sfdefault}{bx}{n}

%% file: sec/0_abstract.tex
\begin{abstract}
    Contrastive learning has become the standard paradigm for multimodal representation learning, aligning semantically matched pairs while separating mismatched ones in a shared embedding space. 
    Although various contrastive training strategies have been proposed across negative selection, batch construction, and gradient modulation, their actual effectiveness under controlled, reproducible conditions remains poorly understood. 
    This motivates a fundamental question: \textit{Do sophisticated contrastive strategies consistently bring meaningful downstream gains in multimodal embedding learning?}
    To address this, we systematically evaluate representative strategies under a unified benchmarking framework on MMEB. 
    Surprisingly, direct evaluations reveal that most existing strategies offer marginal gains over a standard aligned baseline, with several even degrading overall performance. 
    To uncover the underlying failure modes, we introduce an \textbf{MLLM-as-a-Judge} diagnostic framework that dissects top-1 retrieval behaviors into labeled positives, false negatives (unlabeled valid matches), and hard/easy negative errors. 
    Our systematic analysis yields three key insights: 
    (1) \textit{Gradient distortion:} Upweighting gradients from high-similarity negatives consistently hurts downstream generalization; 
    (2) \textit{Distribution illusion:} Wider separation between positive and negative similarity distributions does not necessarily translate to better retrieval performance; and 
    (3) \textit{Synergistic training:} While false-negative mitigation or hard-batch construction yields limited gains in isolation, integrating hard-negative mining, false-negative mitigation, and hard-batch construction achieves optimal performance, delivering an average improvement of \textbf{6.0\%} on MMEB with Qwen2-VL-2B.
    Overall, our findings demystify popular contrastive strategies and provide actionable recipes for training universal multimodal embedding models. Our code and resources are available on \href{https://github.com/alibaba-multimodal-industrial-ai/UME-Recipes}{GitHub} and \href{https://huggingface.co/alibaba-multimodal-industrial-ai/UME-Recipes}{Hugging Face}.
\end{abstract}

%% file: sec/1_intro.tex
\section{Introduction}

Contrastive learning (CL) has emerged as a widely adopted approach for training multimodal embedding models~\citep{radford2021clip,jia2021align}. By pulling semantically matched inputs closer and pushing mismatched ones apart, CL maps heterogeneous multimodal inputs into a shared representation space, supporting various downstream tasks such as classification, retrieval, and visual question answering~\citep{jiang2025vlm2vec,meng2026vlm2vecv2,huang2026mmebv3}. Under this paradigm, early multimodal embedding models primarily adopted CLIP-style dual encoders~\citep{radford2021clip,cherti2023openclip,sun2023evaclip}, which independently encode images and text and align their representations through large-scale contrastive pretraining. Recently, research has shifted toward a new paradigm of universal multimodal embedding models~\citep{jiang2024e5v,zhang2025gme,li2026qwen3vlembedding}, which leverage the instruction-following and multimodal understanding capabilities of multimodal large language models (MLLMs) to accommodate more diverse inputs and tasks. Across both architectural paradigms, different contrastive training strategies have been explored, including gradient modulation~\citep{lan2025llave,xue2025qqmm}, false-negative mitigation~\citep{Zhong2017kreciprocal,balmaseda2025glofnd}, batch construction~\citep{kim2026falcon,dutta2026hobit}, and hard-negative mining~\citep{thirukovalluru2026b3,gu2026unimev2}.

However, despite their reported gains, the actual effectiveness of these strategies remains unclear. Existing methods are evaluated across different model architectures and experimental settings, making it difficult to isolate the effectiveness of training strategies from hyperparameter tuning. This raises two central questions:~\ding{182} how existing training strategies compare in their effectiveness; and~\ding{183} what explains their success or failure.

To answer the two questions, we systematically evaluate representative contrastive training strategies under a unified experimental setting. We conduct all evaluations on the Massive Multimodal Embedding Benchmark (MMEB), a comprehensive benchmark spanning 36 datasets across four meta-task categories~\citep{jiang2025vlm2vec}. Our studies show that most existing contrastive training strategies yield only marginal gains over a well-aligned baseline and many even degrade overall performance. To better understand these strategies, we develop an MLLM-as-a-Judge framework that categorizes top-1 predictions into labeled positives, false negatives (unlabeled yet semantically valid positives), and hard- and easy-negative errors. Our analysis reveals three key findings. 
First, upweighting gradients from high-similarity negatives degrades performance because similarity reflects hardness rather than label reliability. As a result, both hard negatives and false negatives receive larger gradient weights and are pushed farther from the query, while easy negatives contribute relatively less to training. 
Second, wider separation between positive and negative similarity distributions does not reliably indicate better downstream performance, because aggregate distributional separation does not imply that each positive is better separated from its corresponding negatives at the instance level. 
Third, either false-negative mitigation or hard-batch construction alone yields limited gains. Filtering false negatives improves label reliability but can also mistakenly filter out informative hard negatives, whereas constructing hard batches increases the density of valuable in-batch negatives without guaranteeing the semantic validity of individual negatives. Combining instance-level hard-negative mining and false-negative mitigation with batch-level hard-batch construction provides both reliable and informative supervision, leading to the strongest performance. 

Overall, the contributions of this work are threefold:
\begin{itemize}
    \item We systematically compare representative contrastive training strategies under a unified setting on MMEB using Qwen2-VL-2B/7B~\citep{wang2024qwen2vl} and Qwen3.5-2B~\citep{qwen35blog}, showing that most strategies yield no meaningful downstream gains and many even degrade overall performance, revealing a notable discrepancy between their reported gains and actual effectiveness.
    \item We develop an MLLM-as-a-Judge analysis framework that decomposes top-1 predictions into labeled positives, unlabeled yet semantically valid positives, and hard- and easy-negative errors, providing a fine-grained diagnosis of model behavior beyond aggregate benchmark scores.
    \item We uncover three key findings for contrastive training: upweighting gradients from high-similarity negatives degrades performance; separation between positive and negative similarity distributions does not reliably indicate downstream performance; and jointly integrating instance-level hard-negative mining and false-negative mitigation with hard-batch construction yields the strongest performance.
\end{itemize}

%% file: sec/2_related_work.tex
\section{Related Work}

\subsection{Multimodal Embedding Learning}

CLIP~\citep{radford2021clip} established a scalable dual-encoder paradigm for learning transferable image-text embeddings through large-scale contrastive pre-training. Subsequent studies improved its scalability and training efficiency, including OpenCLIP~\citep{cherti2023openclip} with reproducible open-data training, SigLIP~\citep{zhai2023siglip} with a pairwise sigmoid objective, and EVA-CLIP~\citep{sun2023evaclip} with improved initialization and optimization. Nevertheless, CLIP-style dual encoders face several limitations: 1) global image-text alignment provides limited fine-grained spatial grounding~\citep{zhong2022regionclip}; 2) independent modality encoding makes it difficult to generate task-specific representations conditioned on instructions~\citep{jiang2025vlm2vec}; and 3) these models remain limited in understanding compositional relationships, long descriptions, and high-resolution visual details~\citep{ma2023crepe,zhang2024longclip,shi2025scaling}.

To address these issues, recent studies adopt MLLMs as more flexible backbones for multimodal representation learning. E5-V~\citep{jiang2025vlm2vec} adapts an MLLM into a universal embedding model using prompts and text-only contrastive training~\citep{jiang2024e5v}. VLM2Vec~\citep{jiang2025vlm2vec} introduces a general framework for converting MLLMs into instruction-aware embedding models and establishes MMEB, covering 36 datasets across classification, VQA, retrieval, and visual grounding. GME~\citep{zhang2025gme} further expands training to large-scale single-modal, cross-modal, and fused-modal data. More recently, Seed1.6-Embedding~\citep{bytedance2025seedembedding} scales unified representations to text, images, videos, and mixed-modal inputs, while Qwen3-VL-Embedding~\citep{li2026qwen3vlembedding} introduces an open family of embedding and reranking models. Together, these advances demonstrate the potential of MLLMs as flexible and powerful backbones for universal multimodal embedding learning.

\subsection{Contrastive Training Strategies for Multimodal Embeddings}

Building on both CLIP-style dual encoders and MLLM-based embedding models, recent studies have explored improving CL through gradient modulation~\citep{Sun2020circle,lan2025llave}, false-negative mitigation~\citep{Zhong2017kreciprocal,Byun2024mafa,balmaseda2025glofnd}, batch construction~\citep{thirukovalluru2026b3,kim2026falcon}, and hard-negative mining~\citep{robinson2021clwhn,thirukovalluru2026b3}. UniME-V2~\citep{gu2026unimev2} employs an MLLM judge to mine hard negatives, filter false negatives, and generate soft semantic labels. EGA~\citep{xue2025qqmm} explicitly amplifies the gradients associated with hard negatives based on the positive-negative similarity margin. HOBIT~\citep{dutta2026hobit} dynamically selects hard yet non-contradictory in-batch negatives. However, differences in model architectures and experimental settings leave the relative effectiveness of these strategies and the reasons for their success or failure unclear. We therefore benchmark representative strategies under a unified setting and diagnose their behavior beyond aggregate scores to identify genuine methodological gains and provide more reliable guidance for future designs.

%% file: sec/3_exp.tex
\section{Experiments}

\subsection{Contrastive Training Strategies}

\begin{table*}[t]
\centering
\captionsetup{skip=3pt}
\caption{\textbf{\textit{Overview of representative contrastive training strategies.}}
\emph{Dependencies} indicate whether a method requires external scoring or additional preprocessing.
Entries under \emph{Optimization Dimensions} indicate what aspects each contrastive learning method targets explicitly.
\textit{Static} denotes that the corresponding optimization signal is fixed before training, whereas
\textit{Adaptive} denotes that it evolves with the student model.
}
\label{tab:strategy_overview}

\small
\renewcommand{\arraystretch}{1.12}

\begin{tabular*}{\textwidth}{@{\extracolsep{\fill}}lcccccc@{}}
\toprule

&
\multicolumn{2}{c}{\textbf{Dependencies}} &
\multicolumn{4}{c}{\textbf{Optimization Dimensions}} \\
\cmidrule(lr){2-3}
\cmidrule(lr){4-7}

\textbf{Method} &
\makecell{\textbf{External}\\\textbf{Scorer}} &
\makecell{\textbf{Pre-}\\\textbf{processing}} &
\makecell{\textbf{Gradient}\\\textbf{Modulation}} &
\makecell{\textbf{False-Negative}\\\textbf{Mitigation}} &
\makecell{\textbf{Batch}\\\textbf{Construction}} &
\makecell{\textbf{Hard-Negative}\\\textbf{Mining}} \\
\midrule

\rowcolor{groupABody}
LLaVE
& -- & -- & Adaptive & -- & -- & -- \\

\rowcolor{groupABody}
EGA
& -- & -- & Adaptive & -- & -- & -- \\

\rowcolor{groupBBody}
GloFND
& -- & -- & -- & Adaptive & -- & -- \\

\rowcolor{groupBBody}
FN Risk
& -- & \cmark & -- & Adaptive & -- & -- \\

\rowcolor{groupCBody}
FALCON
& -- & \cmark & -- & -- & Adaptive & -- \\

\rowcolor{groupCBody}
HOBIT
& -- & \cmark & -- & -- & Adaptive & -- \\

\rowcolor{groupDBody}
UniME-V2
& \cmark & \cmark & Static & Static & -- & Static \\

\rowcolor{headergray}
B3
& \cmark & \cmark & -- & Static & Static & Static \\

\bottomrule
\end{tabular*}
\end{table*}

We evaluate eight representative strategies that optimize different aspects of contrastive training. As summarized in \Cref{tab:strategy_overview}, we characterize each strategy by its dependencies and optimization dimensions. Dependencies indicate whether a method requires external scoring or additional preprocessing. An \emph{external scorer} refers to a separately pretrained teacher, retriever, or judge that is not jointly updated with the student model, while \emph{preprocessing} denotes the construction of embeddings, rankings, graphs, or candidate pools before training begins. The four optimization dimensions target different aspects of contrastive training. \emph{Gradient modulation} modifies the contrastive objective or sample weights to adjust the gradient contributions of different negative pairs. \emph{False-negative mitigation} identifies and mitigates semantically valid samples that are incorrectly treated as negatives. \emph{Batch construction} organizes training examples to improve the informativeness of in-batch negatives, whereas \emph{hard-negative mining} identifies or introduces challenging negative candidates at the instance level. A strategy is marked as \emph{static} for a given dimension when its supervision or optimization structure remains fixed throughout training, and as \emph{adaptive} when it evolves with the current model, historical representations, or a jointly learned controller. Detailed descriptions of the individual strategies are provided in Appendix~\ref{app:strategy_details}.

\subsection{Basic Setup}
\label{sec:basic_setup}

\begin{table*}[t]
\centering
\caption{\itbf{Training-setting alignment.}
We align the learning rate (LR), maximum number of visual tokens, LoRA rank, and pooling strategy for the VLM2Vec baseline on MMEB using Qwen2-VL-2B-Instruct.
\textit{Cls.}, \textit{VQA}, \textit{Ret.}, and \textit{VG} denote image classification, visual question answering, image-level retrieval, and visual grounding, respectively.
\textit{Overall} denotes the mean Precision@1 across all 36 MMEB evaluation datasets.
The configuration adopted for all subsequent experiments is highlighted in \textbf{bold}.
}
\label{tab:hyperparameter_alignment}

\small
\setlength{\tabcolsep}{4.5pt}
\renewcommand{\arraystretch}{1.08}

\resizebox{\textwidth}{!}{%
\begin{tabular}{ccccccccc}
\toprule

\multicolumn{4}{c}{\textbf{Training Configuration}} &
\multicolumn{5}{c}{\textbf{MMEB}} \\

\cmidrule(lr){1-4}
\cmidrule(lr){5-9}

\textbf{LR} &
\textbf{Max. Visual Tokens} &
\textbf{LoRA Rank} &
\textbf{Pooling Strategies} &
\textbf{Cls.} &
\textbf{VQA} &
\textbf{Ret.} &
\textbf{VG} &
\textbf{Overall} \\

\midrule

% Hyperparameter alignment: very light blue
\rowcolor[rgb]{0.965,0.980,0.992}
2e-5 & 2,880 & 8 & Last Token
& 58.7 & 49.3 & 65.0 & 72.9 & 59.8 \\

\rowcolor[rgb]{0.965,0.980,0.992}
1e-4 & 2,880 & 8 & Last Token
& 60.0 & 56.5 & 67.3 & 76.0 & 63.2 \\

\rowcolor[rgb]{0.965,0.980,0.992}
1e-4 & 1,280 & 8 & Last Token
& 59.7 & 56.5 & 67.3 & 76.1 & 63.2 \\

\rowcolor[rgb]{0.965,0.980,0.992}
1e-4 & 2,880 & 4 & Last Token
& 60.7 & 55.4 & 65.2 & 76.5 & 62.5 \\

\rowcolor[rgb]{0.965,0.980,0.992}
1e-4 & 2,880 & 32 & Last Token
& 61.9 & 57.7 & 69.2 & 76.5 & 64.8 \\

% Adopted configuration: light green
\rowcolor[rgb]{0.925,0.970,0.930}
\textbf{1e-4} &
\textbf{2,880} &
\textbf{16} &
\textbf{Last Token} &
\textbf{63.3} &
\textbf{57.6} &
\textbf{68.6} &
\textbf{77.9} &
\textbf{65.1} \\

\midrule

% Pooling comparison: very light gray
\rowcolor[rgb]{0.970,0.970,0.970}
1e-4 & 2,880 & 16 & Mean
& 46.8 & 44.2 & 40.9 & 32.2 & 42.5 \\

\rowcolor[rgb]{0.970,0.970,0.970}
1e-4 & 2,880 & 16 & \texttt{<embedding>}
& 53.3 & 46.6 & 62.4 & 72.9 & 56.6 \\

\bottomrule
\end{tabular}%
}
\end{table*}

\paragraph{Training Setup.}

Following recent work on MLLM-based multimodal embedding models, we adopt Qwen2-VL-2B-Instruct~\citep{wang2024qwen2vl} as the shared backbone and fine-tune it using LoRA~\citep{hu2021lora}. We use the widely adopted VLM2Vec~\citep{jiang2025vlm2vec}, trained on MMEB using the standard InfoNCE objective with in-batch negatives, as our baseline. Before
comparing different training strategies, we align four key training choices for the baseline: learning rate, maximum number of visual tokens, LoRA rank, and embedding pooling. As shown in \Cref{tab:hyperparameter_alignment}, the best-performing configuration uses a learning rate of $1\times10^{-4}$, up to 2,880 visual tokens, a LoRA rank of 16, and last-token pooling, achieving an overall MMEB score of 65.1. We use this fixed configuration, together with the training data and optimization budget, for all subsequent comparisons. All experiments are conducted on $8\times$ NVIDIA H100 GPUs. 

\paragraph{Evaluation Setup.}

We evaluate all models on the MMEB benchmark using Precision@1 as the primary metric. We report results across MMEB's four meta-task categories: image classification, visual question answering, image-level retrieval, and visual grounding, together with the overall mean across all 36 datasets. All methods use the same task-specific instructions and input templates during training and evaluation, thereby controlling for differences in prompt formatting. Additional backbone and evaluation details are provided in Appendix~\ref{app:model_backbones} and Appendix~\ref{app:benchmark}.

Notably, the aligned baseline already achieves an MMEB score of 65.1, comparable to LLaVE-2B's originally reported score of 65.2~\citep{lan2025llave} and higher than UniME-V2-2B's originally reported score of 63.6~\citep{gu2026unimev2}. This observation highlights the importance of aligning basic training configurations before attributing performance gains to a particular strategy.

\subsection{MMEB Results}

\begin{table*}[t]
\centering
\caption{
\itbf{Comparison of multimodal contrastive training strategies on MMEB using Qwen2-VL-2B.}
The evaluation metric is mean Precision@1 (\%) across the corresponding datasets.
IND and OOD denote the averages over in-distribution and out-of-distribution
datasets, respectively.
GM, FN, HN, and BC denote gradient modulation, false-negative mitigation,
hard-negative mining, and batch construction, respectively.
Training time denotes the training wall-clock time; one-time preprocessing
costs are excluded.
The best results are shown in \textbf{bold}.
}
\label{tab:mmeb_unified_comparison}

\setlength{\tabcolsep}{5.2pt}
\renewcommand{\arraystretch}{1.08}

\resizebox{\textwidth}{!}{%
\begin{tabular}{l c c c c c c c c}
\toprule

\multirow[c]{2}{*}[-1.5ex]{\textbf{Method}}
&
\multirow[c]{2}{*}[-1.5ex]{
    \makecell[c]{\textbf{Training Time}\\\textbf{(h:mm)}}
}
&
\multicolumn{4}{c}{\textbf{Per Meta-Task Score}}
&
\multicolumn{3}{c}{\textbf{Average Score}}
\\

\cmidrule(lr){3-6}
\cmidrule(lr){7-9}

&
&
\makecell[c]{\textbf{Image}\\\textbf{Classification}}
&
\makecell[c]{\textbf{Visual Question}\\\textbf{Answering}}
&
\makecell[c]{\textbf{Image-level}\\\textbf{Retrieval}}
&
\makecell[c]{\textbf{Visual}\\\textbf{Grounding}}
&
\textbf{IND}
&
\textbf{OOD}
&
\textbf{Overall}
\\

\midrule

\multicolumn{2}{l}{\textit{No. of Datasets $\rightarrow$}}
& 10 & 10 & 12 & 4 & 20 & 16 & 36 \\

\midrule

\rowcolor{baselinegray}
VLM2Vec (Baseline)
& 19:12
& 63.3 & 57.6 & 68.6 & 77.9
& 69.8 & 59.2 & 65.1 \\

\midrule

\rowcolor{groupAHead}
\multicolumn{9}{c}{\textit{Baseline + GM}} \\

\midrule

\rowcolor{groupABody}
LLaVE
& 19:24
& 57.8 & 58.4 & 68.9 & 78.0
& 69.3 & 57.2 & 63.9 \\

\rowcolor{groupABody}
EGA
& 19:26
& 63.3 & 57.4 & 68.1 & 77.3
& 69.9 & 58.5 & 64.8 \\

\midrule

\rowcolor{groupBHead}
\multicolumn{9}{c}{\textit{Baseline + FN}} \\

\midrule

\rowcolor{groupBBody}
GloFND
& 15:05
& 62.6 & 55.7 & 68.4 & 77.5
& 68.4 & 59.1 & 64.3 \\

\rowcolor{groupBBody}
FN Risk
& 17:50
& 63.5 & 57.6 & 69.1 & 74.8
& 69.8 & 58.9 & 65.0 \\

\midrule

\rowcolor{groupCHead}
\multicolumn{9}{c}{\textit{Baseline + BC}} \\

\midrule

\rowcolor{groupCBody}
FALCON
& 15:03
& 62.4 & 57.9 & 68.8 & 76.2
& 69.5 & 59.0 & 64.8 \\

\rowcolor{groupCBody}
HOBIT
& 20:06
& 62.2 & 58.1 & 69.6 & 78.3
& 69.7 & 59.9 & 65.3 \\

\midrule

\rowcolor{groupDHead}
\multicolumn{9}{c}{\textit{Baseline + FN + HN + GM}} \\

\midrule

\rowcolor{groupDBody}
UniME-V2
& 39:31
& 62.9 & 57.6 & 69.2 & 76.8
& 68.4 & 60.9 & 65.1 \\

\midrule

\rowcolor{groupEHead}
\multicolumn{9}{c}{\textit{Baseline + FN + HN + BC}} \\

\midrule

\rowcolor{groupEBody}
B3
& 29:44
& \textbf{67.3} & \textbf{63.0}
& \textbf{71.4} & \textbf{81.4}
& \textbf{72.5} & \textbf{64.7}
& \textbf{69.0} \\

\midrule

\rowcolor{groupFHead}
\multicolumn{9}{c}{\textit{Baseline + FN + HN + BC + GM}} \\

\midrule

\rowcolor{groupFBody}
B3 + EGA
& 30:07
& 66.7 & 62.7 & 71.2 & 80.0
& \textbf{72.5} & 63.7 & 68.6 \\

\bottomrule
\end{tabular}%
}
\end{table*}

Using the best-performing configuration identified in Section~\ref{sec:basic_setup}, we train and evaluate all eight strategies, together with an additional B3 + EGA variant, using Qwen2-VL-2B as the backbone. As shown in \Cref{tab:mmeb_unified_comparison}, the gradient-modulation methods LLaVE~\citep{lan2025llave} and EGA~\citep{xue2025qqmm} achieve overall scores of 63.9 and 64.8, respectively. For false-negative mitigation, GloFND~\citep{balmaseda2025glofnd} reaches 64.3, while FN Risk, our adaptive false-negative filtering baseline based on k-reciprocal neighbors~\citep{Zhong2017kreciprocal}, obtains 65.0. Methods targeting either gradient modulation or false-negative mitigation fail to surpass the aligned VLM2Vec baseline of 65.1, indicating that optimizing either dimension in isolation does not improve performance. Batch-construction methods similarly provide no consistent improvement: FALCON~\citep{kim2026falcon} obtains 64.8, while HOBIT~\citep{dutta2026hobit} marginally improves the baseline from 65.1 to 65.3.

When multiple optimization dimensions are combined, UniME-V2~\citep{gu2026unimev2} matches the baseline with an overall score of 65.1. In contrast, B3~\citep{thirukovalluru2026b3} substantially improves the overall score from 65.1 to 69.0, with consistent gains across all four meta-task categories. It also improves the IND and OOD averages by 2.7 and 5.5 points, respectively. Finally, adding EGA to B3 provides no further benefit and reduces the overall score from 69.0 to 68.6. Overall, optimizing individual dimensions yields no consistent gains. Among the evaluated strategies that combine multiple dimensions, only B3 achieves a substantial improvement by jointly optimizing false-negative mitigation, hard-negative mining, and batch construction. Experiments with Qwen2-VL-7B and Qwen3.5-2B show similar trends, as detailed in Appendix~\ref{app:cross_model_comparison}. This divergence motivates a deeper analysis in the following section.

%% file: sec/4_analysis.tex
\section{Understanding the Effects of Contrastive Training Strategies}
\label{sec:understanding}

\subsection{MLLM-as-a-Judge Analysis}
\label{sec:mllm_judge}

Standard Precision@1 treats a top-ranked candidate as correct only when it matches the labeled target, grouping all other outcomes as errors. However, multimodal embedding benchmarks may contain unlabeled but semantically valid matches (i.e., false negatives), while genuinely incorrect predictions may be either plausible but subtly incorrect hard negatives or clearly irrelevant easy negatives. Aggregate scores conflate these distinct outcomes, obscuring how different training strategies affect model behavior.

To obtain a finer-grained diagnosis, we introduce an MLLM-as-a-Judge framework, as illustrated in \Cref{fig:mllm_judge}. The complete prompt template and structured output schema are provided in Appendix~\ref{app:mllm_judge_protocol}. We randomly sample 200 cases from each of the 36 MMEB evaluation datasets and evaluate every method on the same $36 \times 200 = 7{,}200$ cases. Given a query, its designated positive, and the model's top-1 prediction, the judge assigns the prediction to one of four categories: (1) \emph{Labeled Positive (LP)}, which matches the annotated target; (2) \emph{False Negative (FN)}, which differs from the annotated target but remains semantically valid; (3) \emph{Hard Negative (HN)}, which is semantically related and superficially plausible but fails one or more critical query constraints; or (4) \emph{Easy Negative (EN)}, which is clearly irrelevant to the query. We define \emph{Semantic Valid} as the sum of LP and FN, and \emph{Semantic Error} as the sum of HN and EN. This decomposition distinguishes exact target matches from alternative valid predictions and fine-grained semantic errors from clearly irrelevant results, enabling a more precise diagnosis of the gains and failure modes of individual training strategies. We further verify the robustness of this analysis to the evaluation-set size and the choice of MLLM judge in Appendix~\ref{app:robustness_mllm_as_judge}.

\begin{figure*}[t]
    \centering
    \includegraphics[width=\textwidth]{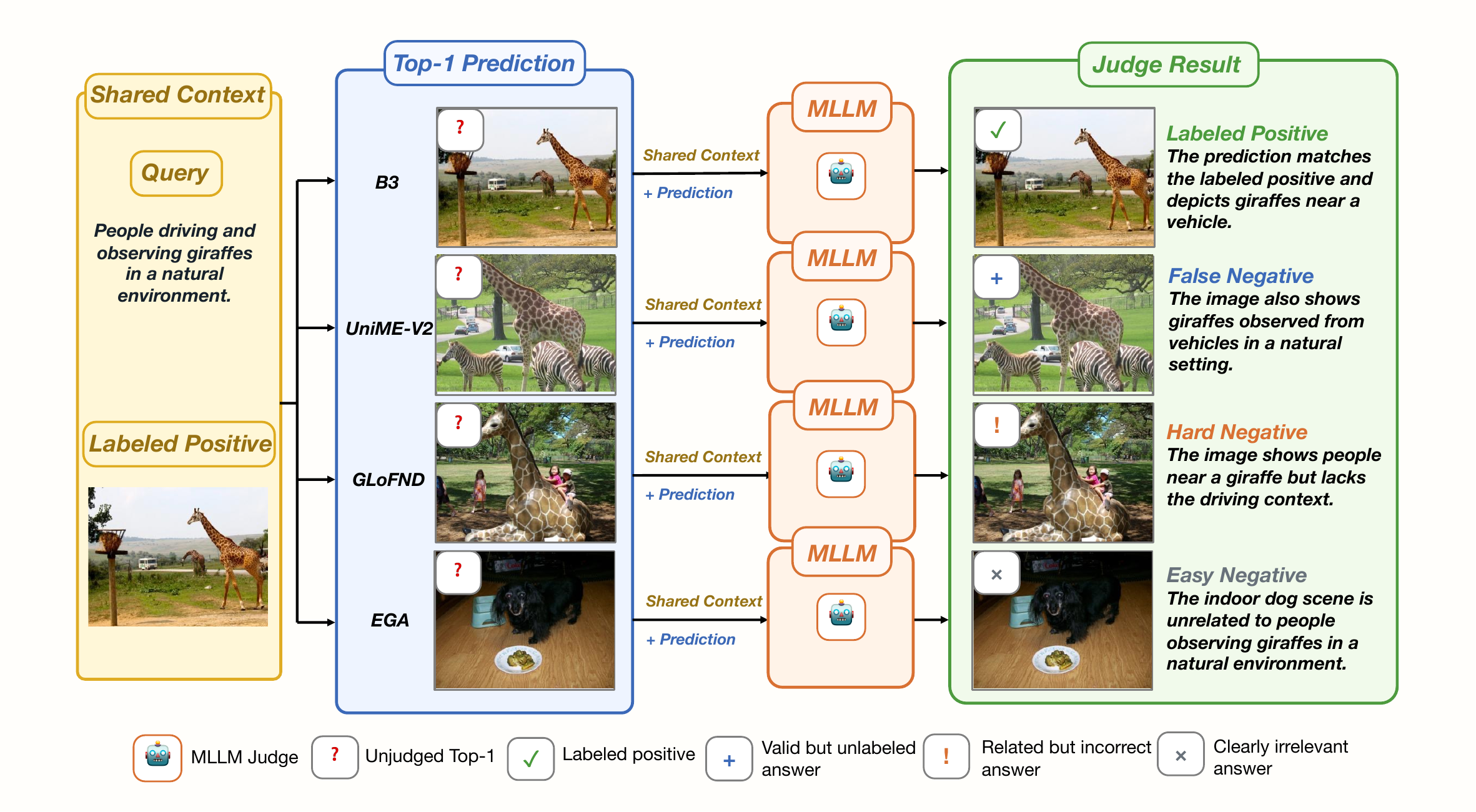}
    \caption{
    \itbf{Overview of the MLLM-as-a-Judge analysis framework.}
    Given the same query and labeled positive, we collect top-1 predictions produced by different contrastive training strategies. The MLLM judge categorizes the prediction as a labeled positive, an alternative valid positive (i.e., false negative), a hard-negative error, or an easy-negative error. This decomposition enables a more fine-grained analysis than standard exact-match Precision@1.
    }
    \label{fig:mllm_judge}
\end{figure*}

\begin{table*}[t]
\centering
\caption{
\itbf{MLLM-as-a-Judge analysis of top-1 predictions on MMEB.}
All methods are evaluated on the same 7,200 cases, with 200 cases sampled
from each of the 36 datasets.
\textit{Labeled Positive} matches the annotated target;
\textit{False Negative} is a different but semantically valid answer;
and \textit{Semantic Valid} is their sum.
\textit{Hard Negative} denotes a related but incorrect prediction, while
\textit{Easy Negative} denotes a clearly irrelevant prediction;
\textit{Semantic Error} is their sum.
GM, FN, HN, and BC denote gradient modulation, false-negative mitigation,
hard-negative mining, and batch construction, respectively.
Arrows indicate the preferred direction, and the best results are shown in
\textbf{bold}.
}
\label{tab:mmeb_judge_analysis}

\small
\setlength{\tabcolsep}{5.5pt}
\renewcommand{\arraystretch}{1.10}

\begin{tabular*}{\textwidth}{
    @{\extracolsep{\fill}}
    l c c c c c c c
}
\toprule

\multirow[c]{2}{*}[-1.5ex]{\textbf{Method}}
&
\multicolumn{3}{c}{\textbf{Positive Outcomes (\%)}}
&
\multicolumn{3}{c}{\textbf{Negative Outcomes (\%)}}
&
\multirow[c]{2}{*}[-1.5ex]{
    \makecell[c]{\textbf{Uncertain}\\\textbf{(\%)}}
}
\\

\cmidrule(lr){2-4}
\cmidrule(lr){5-7}

&
\makecell[c]{\textbf{Labeled}\\\textbf{Positive} $\uparrow$}
&
\makecell[c]{\textbf{False}\\\textbf{Negative} $\uparrow$}
&
\makecell[c]{\textbf{Semantic}\\\textbf{Valid} $\uparrow$}
&
\makecell[c]{\textbf{Hard}\\\textbf{Negative} $\downarrow$}
&
\makecell[c]{\textbf{Easy}\\\textbf{Negative} $\downarrow$}
&
\makecell[c]{\textbf{Semantic}\\\textbf{Error} $\downarrow$}
&
\\

\midrule

\rowcolor{baselinegray}
VLM2Vec
& 67.14 & 8.22 & 75.36
& 17.19 & 6.85 & 24.04 & 0.60 \\

\midrule

\rowcolor{groupAHead}
\multicolumn{8}{c}{\textit{Baseline + GM}} \\

\midrule

\rowcolor{groupABody}
LLaVE
& 65.68 & 8.53 & 74.21
& 17.58 & 7.67 & 25.25 & 0.54 \\

\rowcolor{groupABody}
EGA
& 66.68 & 8.31 & 74.99
& 17.24 & 7.04 & 24.28 & 0.74 \\

\midrule

\rowcolor{groupBHead}
\multicolumn{8}{c}{\textit{Baseline + FN}} \\

\midrule

\rowcolor{groupBBody}
GloFND
& 65.63 & 8.28 & 73.91
& 18.78 & 6.65 & 25.43 & 0.67 \\

\rowcolor{groupBBody}
FN Risk
& 66.61 & 8.60 & 75.21
& 17.65 & 6.64 & 24.29 & 0.50 \\

\midrule

\rowcolor{groupCHead}
\multicolumn{8}{c}{\textit{Baseline + BC}} \\

\midrule

\rowcolor{groupCBody}
FALCON
& 66.78 & 8.43 & 75.21
& 18.14 & 5.97 & 24.11 & 0.68 \\

\rowcolor{groupCBody}
HOBIT
& 66.75 & 8.46 & 75.21
& 17.51 & 6.43 & 23.94 & 0.85 \\

\midrule

\rowcolor{groupDHead}
\multicolumn{8}{c}{\textit{Baseline + FN + HN + GM}} \\

\midrule

\rowcolor{groupDBody}
UniME-V2
& 66.43 & \textbf{9.01} & 75.44
& 17.32 & 6.64 & 23.96 & 0.60 \\

\midrule

\rowcolor{groupEHead}
\multicolumn{8}{c}{\textit{Baseline + FN + HN + BC}} \\

\midrule

\rowcolor{groupEBody}
B3
& \textbf{70.58} & 8.22 & \textbf{78.81}
& \textbf{15.58} & 4.94
& \textbf{20.53} & 0.67 \\

\midrule

\rowcolor{groupFHead}
\multicolumn{8}{c}{\textit{Baseline + FN + HN + BC + GM}} \\

\midrule

\rowcolor{groupFBody}
B3 + EGA
& 69.92 & 8.36 & 78.28
& 16.31 & \textbf{4.65}
& 20.96 & 0.76 \\

\bottomrule
\end{tabular*}
\end{table*}

\subsection{Upweighting High-Similarity Negatives Degrades Performance}

\takeaway{Since similarity reflects negative hardness rather than label reliability, upweighting high-similarity negative pairs assigns larger gradient weights to both genuine hard negatives and false negatives. Such similarity-based gradient reweighting amplifies useful signals from genuine hard negatives and misleading signals from false negatives while diminishing the relative gradient contribution of easy negatives, resulting in overall performance degradation.}

LLaVE~\citep{lan2025llave} and EGA~\citep{xue2025qqmm} use different modulation mechanisms to increase the gradient weights assigned to high-similarity negatives. However, the fine-grained results in \Cref{tab:mmeb_judge_analysis} show that neither method improves over the baseline, consistent with their lower overall MMEB scores in \Cref{tab:mmeb_unified_comparison}. Relative to VLM2Vec~\citep{jiang2025vlm2vec}, both methods show modest increases in FN that only partially offset declines in LP accuracy (LLaVE/EGA: FN $+0.31/+0.09$; LP $-1.46/-0.46$). Meanwhile, both HN and EN errors increase (LLaVE/EGA: HN $+0.39/+0.05$; EN $+0.82/+0.19$), resulting in lower Semantic Valid ($-1.15/-0.37$) and higher Semantic Error ($+1.21/+0.24$). Thus, assigning greater gradient weights to high-similarity negatives does not translate into better hard-negative discrimination.

This degradation persists even when negative quality is explicitly controlled. B3~\citep{thirukovalluru2026b3} combines false-negative treatment, hard-negative mining, and hard-batch construction, achieving the highest Semantic Valid rate of 78.81\%. Compared with B3, adding EGA reduces EN errors but substantially increases HN errors (EN: $-0.29$; HN: $+0.73$), resulting in higher Semantic Error ($+0.43$). Meanwhile, the modest increase in FN does not offset the larger decline in LP (FN: $+0.14$; LP: $-0.66$), leading to lower Semantic Valid ($-0.53$). Thus, gradient upweighting remains detrimental even with stronger instance-level negative selection and batch-level construction.

We attribute these results to a mismatch between negative hardness and label reliability. A high similarity score indicates that a candidate is difficult for the current model, but does not determine whether it is semantically invalid. High-similarity negatives therefore include both genuine hard negatives that provide useful gradients and false negatives that introduce misleading gradients. Weighting candidates solely by similarity amplifies both types of signals while reducing the relative gradient contribution of easy negatives, narrowing the supervision provided across different negative types. The B3 + EGA results further suggest that stronger negative selection alone does not resolve this issue, as similarity-based reweighting can still overemphasize the hardest pairs and increase HN errors. Together, these effects explain why upweighting high-similarity negatives can degrade performance rather than improve negative discrimination.

\subsection{Similarity Separation Does Not Reliably Predict Performance}

\begin{figure*}[t]
    \centering
    \includegraphics[
        width=\textwidth
    ]{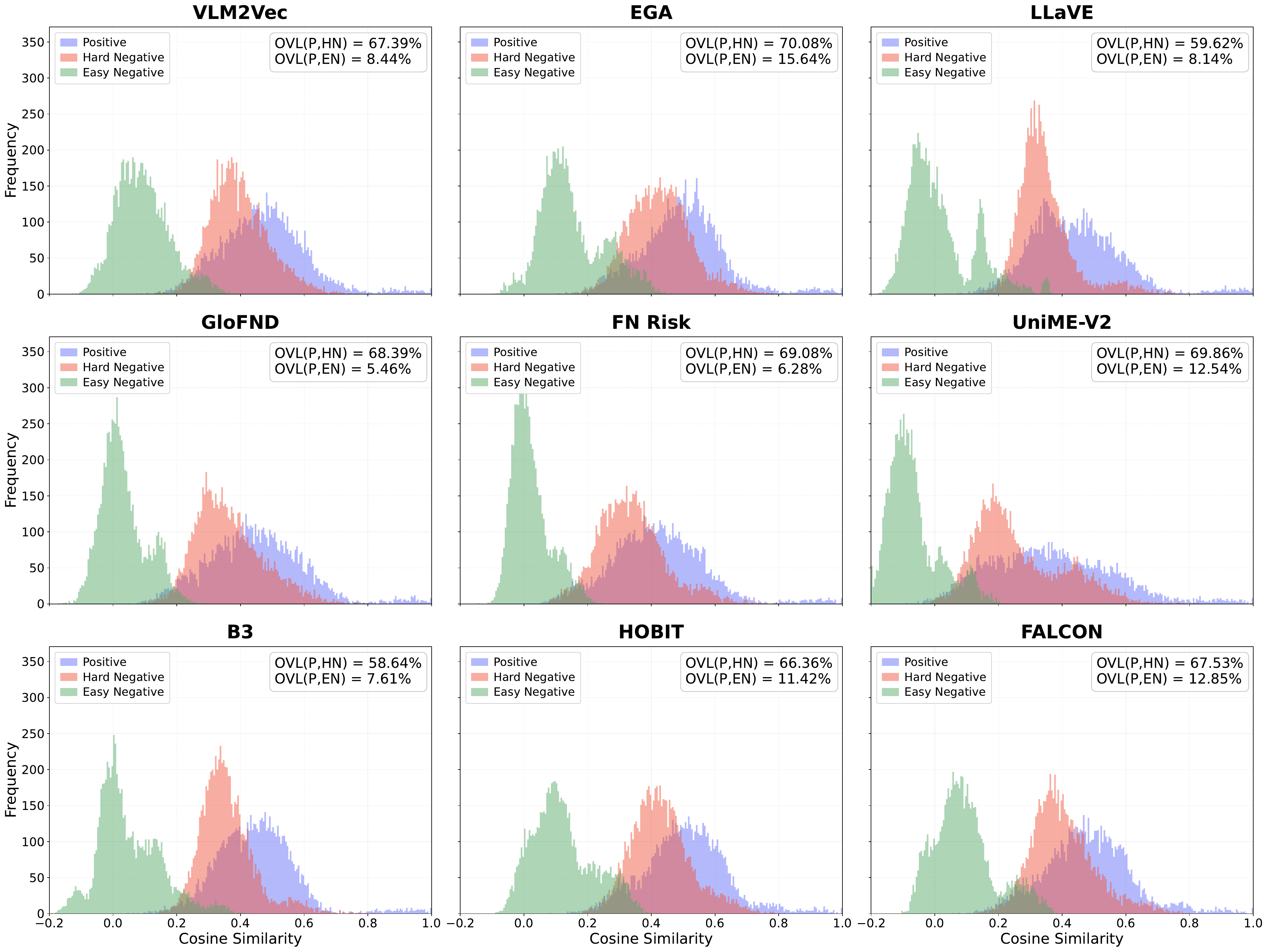}
    \caption{
        \itbf{Similarity distributions of nine training strategies on MMEB.}
        All methods are evaluated on the same 7,200 queries, comprising 200 randomly
        sampled cases from each of the 36 datasets.
        \textit{Positive (P)} denotes the similarity between a query and its designated
        target, while \textit{Hard Negative (HN)} and \textit{Easy Negative (EN)} denote the
        mean similarities to its Top-5 and Bottom-5 non-positive candidates,
        respectively.
        \textit{OVL(P,HN)} and \textit{OVL(P,EN)} measure the distribution overlap between the Positive
        distribution and the HN and EN distributions, respectively; lower values
        indicate better distributional separation.
    }
    \label{fig:mmeb36_embedding_geometry}
\end{figure*}

\takeaway{Since aggregate similarity distributions do not capture the instance-level ordering that determines retrieval accuracy, greater separation between positive and negative distributions does not reliably indicate better downstream performance.}

Prior studies have treated stronger separation between positive and negative similarity distributions as an indicator of embedding quality~\citep{ye2019unsupervised,lan2025llave}. However, \Cref{fig:mmeb36_embedding_geometry} shows that distributional changes do not consistently align with downstream performance. Compared with the baseline, LLaVE reduces both positive--hard-negative and positive--easy-negative overlap (LLaVE: $-7.77/-0.30$), whereas EGA increases both ($+2.69/+7.20$). Despite these opposite changes, both methods degrade consistently in the overall MMEB evaluation (LLaVE/EGA: $-1.2/-0.3$) and the MLLM-as-a-Judge analysis (Semantic Valid: $-1.15/-0.37$; Semantic Error: $+1.21/+0.24$). Thus, distinct similarity distributions can yield the same downstream performance trend.

Comparisons across other strategies further confirm this mismatch. B3 and LLaVE exhibit similar overlap rates (B3/LLaVE: HN $58.64/59.62$; EN $7.61/8.14$), yet B3 outperforms LLaVE by 5.1 MMEB points and 4.60\% in Semantic Valid. Conversely, UniME-V2~\citep{gu2026unimev2} increases both HN and EN overlap relative to the baseline ($+2.47/+4.10$) while retaining the same MMEB score of 65.1 and nearly identical Semantic Valid ($+0.08$). These results show that neither lower distributional overlap nor aggregate similarity geometry reliably predicts downstream performance.

This discrepancy arises because clearer separation between global similarity distributions does not necessarily indicate better positive–negative separation for each query. Globally distinct distributions may still hide many local ranking errors, whereas overlapping distributions can still preserve the correct ordering for most queries. Similarity distributions thus provide insight into embedding geometry, but are not reliable standalone indicators or optimization signals for negative quality.

\subsection{Reliable Contrastive Training Requires Joint Design}
\label{sec:takeaway3}

\takeaway{Neither false-negative mitigation, hard-negative mining, nor hard-batch construction yields consistent gains in isolation. False-negative mitigation removes unreliable negatives without providing informative ones; hard-negative mining incorrectly selects semantically valid false negatives; and batch-level construction does not guarantee sufficiently hard negatives for each instance. The strongest performance comes from jointly filtering false negatives, mining informative instance-level hard negatives, and increasing their coverage through hard-batch construction.}

As shown in \Cref{tab:mmeb_judge_analysis}, false-negative treatment alone provides only limited improvement in recovering semantically valid alternatives, while introducing more errors on LP and HN. Compared with the baseline, GloFND~\citep{balmaseda2025glofnd} and FN Risk~\citep{Zhong2017kreciprocal} improve FN recognition by only $+0.05/+0.38$\%, but decrease LP accuracy by $-1.51/-0.53$ and increase HN errors by $+1.59/+0.46$. Consequently, their Semantic Valid rates decrease by $-1.46/-0.15$\%, consistent with the drops in their overall MMEB scores (GloFND/FN Risk: $-0.8/-0.1$ points). Batch-construction methods provide only limited improvement in recognizing FN and filtering EN, while introducing more errors on LP and HN. Compared with the baseline, FALCON~\citep{kim2026falcon} and HOBIT~\citep{dutta2026hobit} slightly improve FN recognition ($+0.21/+0.24$\%) and reduce EN errors ($-0.88/-0.42$), but simultaneously decrease LP accuracy ($-0.36/-0.39$) and increase HN errors ($+0.95/+0.32$). As a result, their Semantic Valid rates both reach 75.21\%, remaining close to the baseline of 75.36\%, which is consistent with the limited changes in their overall MMEB scores (FALCON/HOBIT: $-0.3/+0.2$ points). These results indicate that addressing either label noise or batch difficulty alone does not consistently improve semantic discrimination.

In contrast, B3 jointly combines false-negative treatment, hard-negative mining, and hard-batch construction. It increases Semantic Valid from 75.36\% to 78.81\% while reducing both HN and EN errors from 17.19\% and 6.85\% to 15.58\% and 4.94\%, respectively. These improvements translate into an overall MMEB score of 69.0, substantially outperforming all isolated strategies. UniME-V2, which combines false-negative treatment and hard-negative mining without hard-batch construction, remains at the baseline score of 65.1, further highlighting the importance of coordinating instance- and batch-level interventions. Moreover, the B3 ablations in Appendix~\ref{app:b3_component_ablation} show that removing any of false-negative treatment, hard-negative mining, or hard-batch construction reduces performance, confirming that the three components provide complementary benefits.

This complementarity arises because the three components address different limitations of contrastive supervision. False-negative treatment improves label reliability but may also reduce training difficulty if informative negatives are removed without replacement. Hard-negative mining supplies challenging pairwise supervision, but requires false-negative filtering to ensure that the selected candidates are semantically invalid. Hard-batch construction then increases the density and coverage of these reliable hard negatives within each optimization step. Combining the three therefore provides supervision that is simultaneously reliable, informative, and sufficiently challenging, leading to the strongest overall performance.

%% file: sec/5_conclusion.tex
\section{Conclusion}

In this work, we systematically examined representative contrastive training strategies for universal multimodal embedding learning under a unified experimental setting on MMEB. Across Qwen2-VL-2B/7B and Qwen3.5-2B, most strategies provide little improvement over a well-aligned baseline and many even degrade performance, revealing that previously reported gains do not persist under controlled comparisons. To understand these results beyond aggregate benchmark scores, we introduced an MLLM-as-a-Judge framework that distinguishes among labeled positives, false negatives, and hard- and easy-negative errors Our analysis reveals that upweighting high-similarity negatives can amplify both useful and misleading gradients. Wider separation between aggregate similarity distributions does not reliably indicate better instance-level retrieval. False-negative mitigation, hard-negative mining, and hard-batch construction address complementary aspects of contrastive supervision and achieve the strongest performance when jointly applied. Overall, these findings clarify the effectiveness and limitations of existing training strategies and provide practical guidance for  training universal multimodal embedding models.

%% file: sec/6_appendix.tex
\begin{center}
    {\LARGE\bfseries Appendix}
\end{center}

\section{Experimental Details}
\label{app:exp_setting}

\subsection{Contrastive Training Strategies}
\label{app:strategy_details}

\paragraph{VLM2Vec.}
VLM2Vec~\citep{jiang2025vlm2vec} provides a general framework for adapting vision-language models into instruction-aware multimodal embedding models through contrastive learning. It encodes multimodal queries and candidates with a shared backbone and optimizes an InfoNCE objective using other candidates in the mini-batch as negatives. We adopt VLM2Vec as the common baseline upon which all evaluated contrastive training strategies are implemented.

\paragraph{LLaVE.}
LLaVE~\citep{lan2025llave} introduces hardness-weighted contrastive learning to emphasize difficult in-batch negatives. It estimates negative hardness using the current query--negative similarity and assigns larger weights to negatives with higher similarity when computing the contrastive objective.

\paragraph{EGA.}
EGA~\citep{xue2025qqmm} analyzes the gradients of the InfoNCE objective and explicitly redistributes gradient contributions among negative samples. It measures hardness using the similarity margin between each negative and the labeled positive, allocating a larger proportion of the negative gradient to samples closer to the positive.

\paragraph{GloFND.}
GloFND~\citep{balmaseda2025glofnd} dynamically estimates a global false-negative threshold for each training instance. The thresholds are updated from sampled mini-batches to approximate per-instance similarity quantiles over the full dataset, and negatives exceeding the corresponding thresholds are excluded from the contrastive objective.

\paragraph{FN Risk.}
FN Risk is our controlled false-negative treatment strategy based on dynamically updated neighborhood information~\citep{Zhong2017kreciprocal}. It estimates the risk of each negative using its similarity to the labeled positive, neighborhood overlap, and mutual-neighbor relationships, and continuously down-weights high-risk negatives instead of removing them entirely.

\paragraph{FALCON.}
FALCON~\citep{kim2026falcon} learns an adaptive scheduler for constructing mini-batches that balance hard-negative informativeness against false-negative risk. For each anchor, the scheduler predicts a hardness quantile over the current similarity distribution and sequentially selects samples at the corresponding quantiles to form a mini-batch.

\paragraph{HOBIT.}
HOBIT~\citep{dutta2026hobit} constructs informative mini-batches using an InfoNCE-motivated criterion that favors negatives similar to the query but dissimilar to its labeled positive. It formulates batch construction as a monotone submodular optimization problem and greedily selects examples that maximize the collective hardness of in-batch negatives.

\paragraph{UniME-V2.}
UniME-V2~\citep{gu2026unimev2} first performs global retrieval to construct candidate pools and then employs an MLLM judge to estimate the semantic alignment of query--candidate pairs. The resulting scores are used to filter false negatives, select diverse hard negatives, and provide soft semantic targets for training the embedding model.

\paragraph{B3.}
B3~\citep{thirukovalluru2026b3} uses a pretrained teacher model to rank training examples and construct a sparse preference graph of hard-negative relationships. It excludes the highest-ranked candidates as potential false negatives, applies community detection to identify groups of mutually hard negatives, and forms mini-batches by sampling from the resulting communities.

\subsection{Model Backbones}
\label{app:model_backbones}

\paragraph{Qwen2-VL.}
Qwen2-VL~\citep{wang2024qwen2vl} is a family of MLLMs that integrates a vision encoder with a decoder-only language model. It adopts Naive Dynamic Resolution to represent images of different resolutions and aspect ratios with variable numbers of visual tokens, together with Multimodal Rotary Position Embedding (M-RoPE) to model one-dimensional text, two-dimensional images, and three-dimensional videos. We use the instruction-tuned 2B model as our primary backbone for controlled comparisons and the 7B model to examine whether our findings remain consistent at a larger model scale.

\paragraph{Qwen3.5.}
Qwen3.5~\citep{qwen35blog} is a native multimodal model jointly pretrained on visual and textual inputs. It combines a vision encoder with a hybrid language backbone that interleaves Gated DeltaNet and gated full-attention layers. We use the 2B model to examine whether our conclusions generalize across model generations and sequence-modeling architectures.

\subsection{Evaluation Benchmark}
\label{app:benchmark}

\paragraph{MMEB.}
The MMEB~\citep{jiang2025vlm2vec} evaluates universal multimodal embedding models on 36 datasets spanning four meta-tasks: image classification, visual question answering, image-level retrieval, and visual grounding. All tasks are formulated as retrieval problems over multimodal candidate sets, and models generate instruction-conditioned embeddings for both queries and candidates. We report Precision@1 averaged across the 36 datasets, together with scores for each meta-task and the in- and out-of-distribution subsets.

\paragraph{Prompt Formatting.}
We follow the official MMEB instruction format without additional prompt
tuning. Queries are encoded using the task-specific instructions provided
by each MMEB dataset. For textual targets, we use modality-level templates:
\texttt{Represent the class label:} for classification,
\texttt{Represent the answer:} for visual question answering, and
\texttt{Represent the image caption:} for image-to-text retrieval.
Other targets, including full images and cropped regions, retain their
original MMEB instructions. The same prompt formatting is used across all
models, training strategies, checkpoints, and ablation experiments.

\section{Ablation Studies}

\begin{table*}[t]
\centering
\caption{
\itbf{Comparison of multimodal contrastive training strategies on MMEB using Qwen3.5-2B and Qwen2-VL-7B.}
We extend the unified comparison in Table~\ref{tab:mmeb_unified_comparison} under the same training and evaluation protocol.
Scores are mean Precision@1 (\%) over the corresponding datasets, and
\textit{Overall} is the average across all 36 MMEB datasets.
GM, FN, HN, and BC denote gradient modulation, false-negative mitigation,
hard-negative mining, and batch construction, respectively.
The best result in each column under each backbone is highlighted in \textbf{bold}.
}
\label{tab:mmeb_backbone_robustness}

\small
\setlength{\tabcolsep}{5.5pt}
\renewcommand{\arraystretch}{1.10}

\resizebox{\textwidth}{!}{%
\begin{tabular}{@{}l l c c c c c@{}}
\toprule

\multirow[c]{2}{*}[-1.5ex]{\textbf{Method}}
&
\multirow[c]{2}{*}[-1.5ex]{
    \makecell[l]{\textbf{Optimization}\\\textbf{Dimensions}}
}
&
\multicolumn{4}{c}{\textbf{Per Meta-Task Score}}
&
\multirow[c]{2}{*}[-1.5ex]{
    \makecell[c]{\textbf{Overall}\\\textbf{Score}}
}
\\

\cmidrule(lr){3-6}

&
&
\makecell[c]{\textbf{Image}\\\textbf{Classification}}
&
\makecell[c]{\textbf{Visual Question}\\\textbf{Answering}}
&
\makecell[c]{\textbf{Image-level}\\\textbf{Retrieval}}
&
\makecell[c]{\textbf{Visual}\\\textbf{Grounding}}
&
\\

\midrule

\multicolumn{2}{@{}l}{\textit{No. of Datasets $\rightarrow$}}
& 10 & 10 & 12 & 4 & 36 \\

\midrule

% =========================================================
% Qwen3.5-2B
% =========================================================
\rowcolor{backbonegray}
\multicolumn{7}{c}{\textbf{\textit{Qwen3.5-2B}}} \\

\midrule

\rowcolor{baselinegray}
VLM2Vec (Baseline)
& --
& 65.2 & 67.4 & 67.3 & 84.6 & 68.7 \\

\rowcolor{groupABody}
EGA
& GM
& 65.4 & \textbf{67.8} & 67.4 & 82.5 & 68.6 \\

\rowcolor{groupBBody}
FN Risk
& FN
& 65.3 & 66.0 & 67.0 & 84.3 & 68.2 \\

\rowcolor{groupCBody}
HOBIT
& BC
& 65.8 & 66.7 & 68.7 & 83.7 & 69.0 \\

\rowcolor{groupDBody}
UniME-V2
& FN + HN + GM
& 63.5 & 63.9 & 65.0 & 84.9 & 66.5 \\

\rowcolor{groupEBody}
B3
& FN + HN + BC
& \textbf{66.9} & 67.2 & \textbf{70.0}
& 86.5 & \textbf{70.2} \\

\rowcolor{groupFBody}
B3 + EGA
& FN + HN + BC + GM
& 66.1 & 67.2 & 69.9 & \textbf{87.2} & 70.0 \\

\midrule

% =========================================================
% Qwen2-VL-7B
% =========================================================
\rowcolor{backbonegray}
\multicolumn{7}{c}{\textbf{\textit{Qwen2-VL-7B}}} \\

\midrule

\rowcolor{baselinegray}
VLM2Vec (Baseline)
& --
& 67.4 & 64.9 & 71.7 & 83.7 & 70.0 \\

\rowcolor{groupABody}
EGA
& GM
& 67.0 & 63.8 & 72.4 & 82.0 & 69.6 \\

\rowcolor{groupBBody}
FN Risk
& FN
& 67.5 & 64.1 & 73.1 & 82.4 & 70.1 \\

\rowcolor{groupCBody}
HOBIT
& BC
& 65.5 & 65.2 & 72.7 & 80.8 & 69.5 \\

\rowcolor{groupDBody}
UniME-V2
& FN + HN + GM
& 64.2 & 65.3 & 73.0 & 81.7 & 69.4 \\

\rowcolor{groupEBody}
B3
& FN + HN + BC
& \textbf{70.6} & \textbf{69.1} & \textbf{75.1}
& \textbf{85.6} & \textbf{73.3} \\

\rowcolor{groupFBody}
B3 + EGA
& FN + HN + BC + GM
& 69.9 & 68.4 & 74.6 & 84.8 & 72.7 \\

\bottomrule
\end{tabular}%
}
\end{table*}

\subsection{Cross-Model Comparison}
\label{app:cross_model_comparison}

To examine whether our observations depend on the Qwen2-VL-2B backbone, we repeat the unified comparison using Qwen3.5-2B and Qwen2-VL-7B. As shown in Table~\ref{tab:mmeb_backbone_robustness}, the results closely reproduce the trends in Table~\ref{tab:mmeb_unified_comparison}. Methods targeting a single optimization dimension provide limited or inconsistent improvements: EGA slightly underperforms VLM2Vec on both additional backbones, while FN Risk and HOBIT yield mixed results. UniME-V2 also fails to consistently improve over the baseline despite combining false-negative mitigation, hard-negative mining, and gradient modulation. In contrast, B3 achieves the strongest overall performance, improving over VLM2Vec by 1.5 points on Qwen3.5-2B and 3.3 points on Qwen2-VL-7B, with improvements across most meta-tasks. Adding EGA to B3 slightly reduces the overall score by 0.2 and 0.6 points, respectively, consistent with the observations on Qwen2-VL-2B. These results demonstrate that our main conclusions generalize across model backbones: jointly improving false-negative reliability, instance-level hard-negative quality, and hard-negative coverage is consistently more effective than optimizing individual components in isolation, while additional gradient modulation provides no complementary benefit in this setting.

\subsection{Robustness of MLLM-as-Judge}
\label{app:robustness_mllm_as_judge}

\begin{table*}[t]
\centering
\caption{
\itbf{Robustness of MLLM-as-Judge analysis.}
We evaluate predictions from the same VLM2Vec-2B checkpoint using
different MLLM judges and evaluation sets.
\textit{Full set} includes all available evaluation cases from the
36 MMEB datasets, while the \textit{Subset} contains the 7,200 cases
used in the main analysis, with 200 cases randomly sampled from each
dataset.
Outcome definitions follow
Table~\ref{tab:mmeb_judge_analysis}.
All values are percentages.
}
\label{tab:judge_scale_model_ablation}

\small
\setlength{\tabcolsep}{5.0pt}
\setlength{\extrarowheight}{1pt}
\renewcommand{\arraystretch}{1.12}

\resizebox{\textwidth}{!}{%
\begin{tabular}{@{}l c c c c c c c@{}}
\toprule

&
\multicolumn{3}{c}{\textbf{Positive Outcomes (\%)}}
&
\multicolumn{3}{c}{\textbf{Negative Outcomes (\%)}}
&
\\

\cmidrule(lr){2-4}
\cmidrule(lr){5-7}

\shortstack[c]{\textbf{Judge /}\\\textbf{Set}}
&
\shortstack[c]{\textbf{Labeled}\\\textbf{Positive} $\uparrow$}
&
\shortstack[c]{\textbf{False}\\\textbf{Negative} $\uparrow$}
&
\shortstack[c]{\textbf{Semantic}\\\textbf{Valid} $\uparrow$}
&
\shortstack[c]{\textbf{Hard}\\\textbf{Negative} $\downarrow$}
&
\shortstack[c]{\textbf{Easy}\\\textbf{Negative} $\downarrow$}
&
\shortstack[c]{\textbf{Semantic}\\\textbf{Error} $\downarrow$}
&
\shortstack[c]{\textbf{Uncertain}\\\textbf{(\%)}}
\\

\midrule

\rowcolor{judgeA}
\shortstack[l]{Qwen3.8-Flash\\\textit{Full set}}
& 67.35
& 8.08
& 75.43
& 17.16
& 6.75
& 23.91
& 0.66
\\

\midrule

\rowcolor{judgeB}
\shortstack[l]{Qwen3.8-Flash\\\textit{Subset}}
& 67.14
& 8.22
& 75.36
& 17.19
& 6.85
& 24.04
& 0.60
\\

\midrule

\rowcolor{judgeC}
\shortstack[l]{GPT-5.4\\\textit{Subset}}
& 67.14
& 8.07
& 75.21
& 16.93
& 6.71
& 23.64
& 1.15
\\

\bottomrule
\end{tabular}%
}
\end{table*}

\begin{table*}[t]
\centering
\caption{
\itbf{Ablation study of B3 using Qwen2-VL-2B on MMEB.}
We remove false-negative mitigation (FN), hard-negative mining (HN), or
batch construction (BC) from the full B3 framework while keeping the remaining
training and evaluation settings unchanged.
Scores are mean Precision@1 (\%) over the corresponding datasets, and
\textit{Overall} is the average across all 36 MMEB datasets.
The best result in each column is highlighted in \textbf{bold}.
}
\label{tab:b3_component_ablation}

\small
\setlength{\tabcolsep}{5.5pt}
\renewcommand{\arraystretch}{1.10}

\resizebox{\textwidth}{!}{%
\begin{tabular}{@{}l l c c c c c@{}}
\toprule

\multirow[c]{2}{*}[-1.5ex]{\textbf{Method}}
&
\multirow[c]{2}{*}[-1.5ex]{
\makecell[l]{\textbf{Optimization}\\\textbf{Dimensions}}
}
&
\multicolumn{4}{c}{\textbf{Per Meta-Task Score}}
&
\multirow[c]{2}{*}[-1.5ex]{
\makecell[c]{\textbf{Overall}\\\textbf{Score}}
}
\\

\cmidrule(lr){3-6}

&
&
\makecell[c]{\textbf{Image}\\\textbf{Classification}}
&
\makecell[c]{\textbf{Visual Question}\\\textbf{Answering}}
&
\makecell[c]{\textbf{Image-level}\\\textbf{Retrieval}}
&
\makecell[c]{\textbf{Visual}\\\textbf{Grounding}}
&
\\

\midrule

\multicolumn{2}{@{}l}{\textit{No. of Datasets $\rightarrow$}}
& 10 & 10 & 12 & 4 & 36 \\

\midrule

\rowcolor{backbonegray}
\multicolumn{7}{c}{\textbf{\textit{Qwen2-VL-2B}}} \\

\midrule

\rowcolor{groupEBody}
B3 (Full)
& FN + HN + BC
& 67.3
& 63.0
& \textbf{71.4}
& \textbf{81.4}
& \textbf{69.0} \\

\rowcolor{ablationA}
Random Batch
& FN + HN
& 67.3
& 60.7
& 71.0
& 78.1
& 67.9 \\

\rowcolor{ablationB}
w/o FN Filtering
& HN + BC
& 67.3
& 61.7
& 70.7
& 80.7
& 68.4 \\

\rowcolor{ablationC}
w/o HN Mining
& FN + BC
& \textbf{67.5}
& \textbf{63.2}
& 71.0
& 79.0
& 68.8 \\

\bottomrule
\end{tabular}%
}
\end{table*}

To examine whether our MLLM-as-Judge analysis is sensitive to the evaluation-set size or the choice of judge, we repeat the analysis on predictions from the same VLM2Vec-2B checkpoint under three configurations (Table~\ref{tab:judge_scale_model_ablation}). Reducing the Qwen3.8-Flash evaluation from the full 36,000 cases to the 7,200-case subset changes every outcome by at most 0.21 percentage points. Replacing Qwen3.8-Flash with GPT-5.4 on the same subset produces similarly small differences: all positive and negative outcome categories change by at most 0.40 percentage points. These results confirm that the 7,200-case evaluation used in the main text is representative of the full evaluation set and yields consistent semantic-error patterns across different judge models.

\subsection{Ablation of the Joint Design in B3}
\label{app:b3_component_ablation}

In Section~\ref{sec:takeaway3}, we demonstrate the effectiveness of the joint design by comparing methods that cover different optimization dimensions. To further identify the contribution of each optimization dimension, we conduct a controlled ablation within B3 by individually removing false-negative mitigation, hard-negative mining, or hard-batch construction while keeping all other implementation and training settings unchanged. As shown in \Cref{tab:b3_component_ablation}, replacing hard-batch construction with random batching reduces the overall MMEB score from 69.0 to 67.9, while removing false-negative filtering or hard-negative mining lowers it to 68.4 and 68.8, respectively. Although some ablations yield marginal improvements on individual meta-tasks, none matches the overall performance of the complete B3 configuration. These results further support our finding: false-negative mitigation, hard-negative mining, and hard-batch construction provide complementary benefits and achieve the strongest performance when optimized jointly.

\section{MLLM-as-Judge Evaluation Protocol}
\label{app:mllm_judge_protocol}

\begin{figure*}[t]
    \centering
    \includegraphics[
        width=\textwidth
    ]{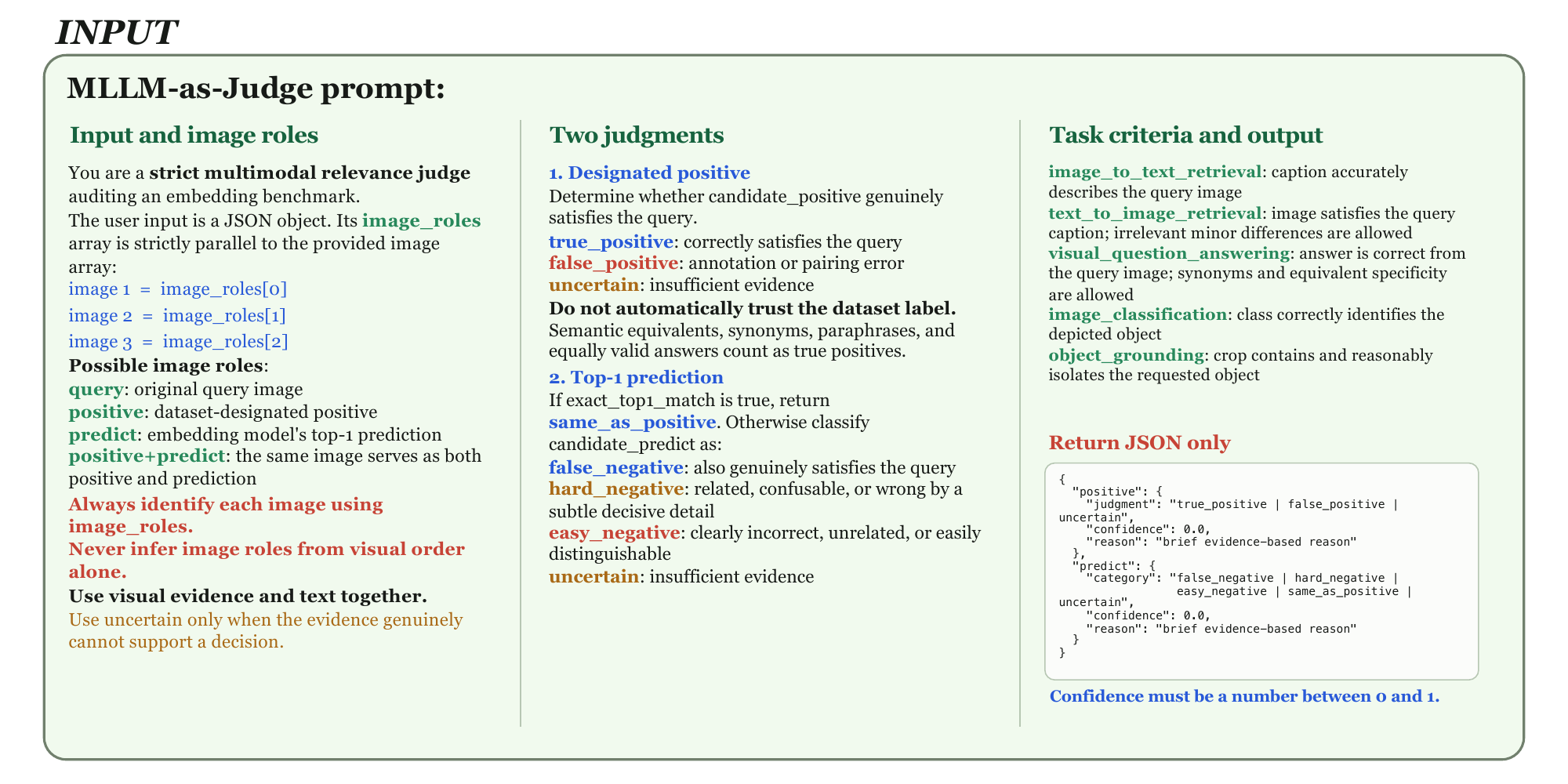}
    \caption{
        \itbf{Overview of the MLLM-as-Judge prompt.}
        The judge receives the query, the dataset-designated positive, the
        model's top-1 prediction, and an explicit mapping between images and their roles. It first verifies the designated positive and then categorizes the top-1 prediction as the designated positive, a false negative, a hard negative, an easy negative, or an uncertain case. Task-specific criteria guide judgments across the five MMEB task families, and a constrained JSON schema enables automatic parsing and aggregation.
    }
    \label{fig:mllm_judge_prompt}
\end{figure*}

Figure~\ref{fig:mllm_judge_prompt} summarizes our MLLM-as-Judge prompt. For each case, the judge receives the query, dataset-designated positive, model's top-1 prediction, task type, and explicit image-role mappings. It first verifies whether the designated positive is valid and then categorizes the top-1 prediction as the designated positive, a false negative, a hard negative, an easy negative, or an uncertain case. Task-specific criteria are provided for the five MMEB task families, and the judgment is returned using a constrained JSON schema for automatic parsing. We use the same prompt and output schema for all methods, with Qwen3.8-Flash as the default judge.